\documentclass{article}
\usepackage[a4paper, total={6in, 9in}]{geometry}
\usepackage[utf8]{inputenc}
\usepackage{url}

\title{AandP: Utilizing Prolog for converting between active sentence and passive sentence with three-steps conversion}

\author{\textbf{Trung Q. Tran}}
\date{}

\usepackage[square,numbers]{natbib}
\usepackage{graphicx}
\usepackage{subcaption}
\usepackage{courier}

\usepackage{listings}
\usepackage{courier}
\usepackage{xcolor}
\usepackage[hidelinks]{hyperref}
\hypersetup{
  colorlinks   = true, 
  urlcolor     = blue, 
  linkcolor    = blue, 
  citecolor   = red 
}

\begin{document}

\maketitle

\begin{abstract}
    I introduce a simple but efficient method to solve one of the critical aspects of English grammar which is the relationship between active sentence and passive sentence. In fact, an active sentence and its corresponding passive sentence express the same meaning, but their structure is different. I utilized Prolog \cite{pereira2002prolog} along with Definite Clause Grammars (DCG) \cite{pereira1980definite} for doing the conversion between active sentence and passive sentence. Some advanced techniques were also used such as Extra Arguments, Extra Goals, Lexicon, etc. I tried to solve a variety of cases of active and passive sentences such as 12 English tenses, modal verbs, negative form, etc. More details and my contributions will be presented in the following sections. The source code is available at \url{https://github.com/tqtrunghnvn/ActiveAndPassive}.
\end{abstract}

\section{Introduction}
\begin{figure}[h!]
\centering
\includegraphics[scale=0.5]{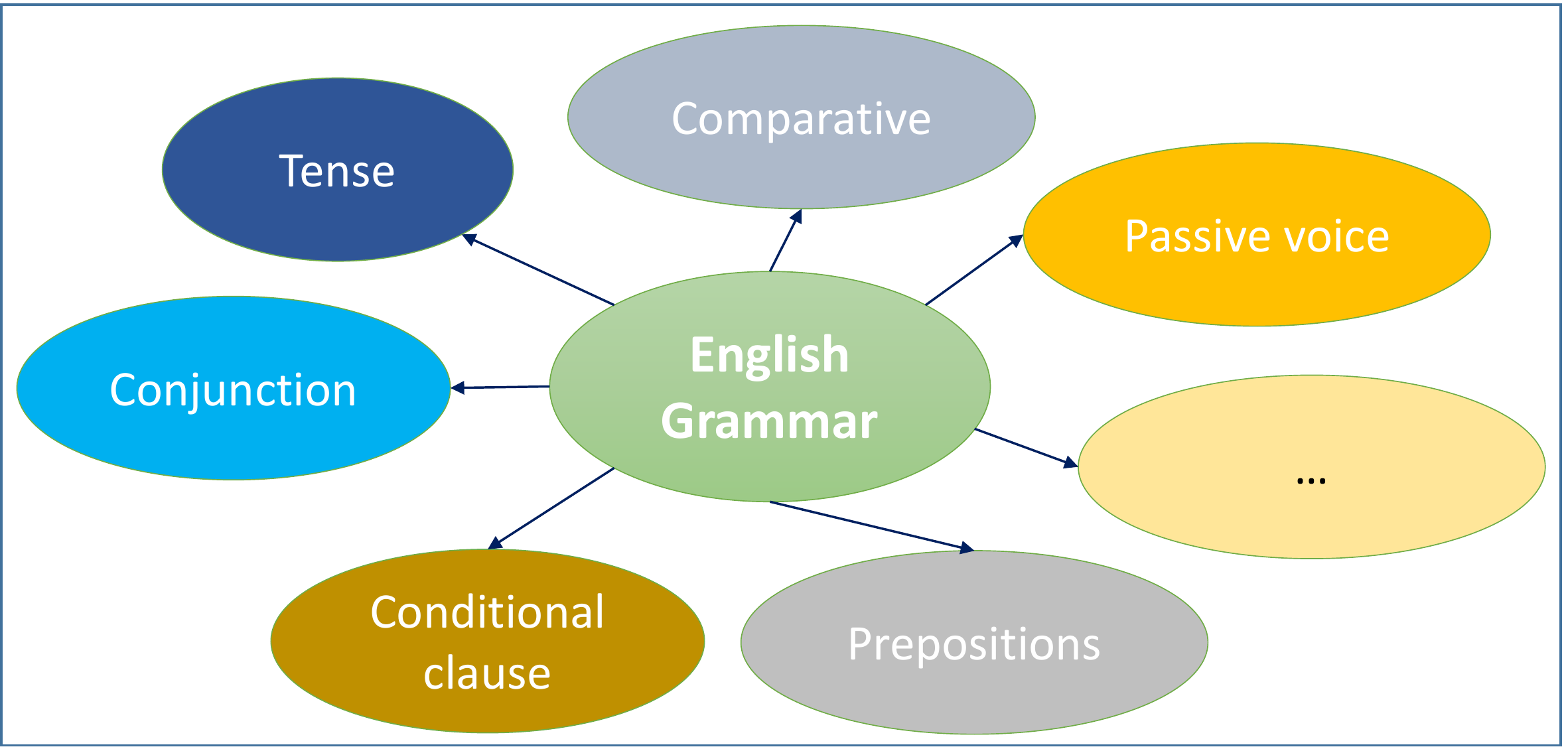}
\caption{A variety of stuff in English grammar}
\label{fig:englishgrammar}
\end{figure}

Language plays a vital role in the human life. A language is a structured system of
communication \cite{davidson2012semantics}. There are various language systems in the world with the estimated number being between 5,000 and 7,000 \cite{wikipedialanguage}. Natural Language Processing (NLP) which we commonly hear is a subfield of linguistics. NLP aims to provide interactions between computers and human languages. The performance of NLP is evaluated by how computers can process and analyze large amounts of natural language data \cite{wikipedianlp}. In terms of language processing, we cannot but mention Computational Linguistics \cite{grishman1986computational}. Computational Linguistics is the scientific study of language from a computational perspective, and thus an interdisciplinary field, involving linguistics, computer science, mathematics, logic, cognitive science, and cognitive psychology.

One of the most useful tools for studying computational linguistics is Prolog programming language \cite{pereira2002prolog}. Prolog is a logic programming language associated with artificial intelligence and computational linguistics. Prolog can help deal with issues related to not only logic puzzle (Cryptoarithmetic puzzles, Zebra Puzzle, etc.) but also natural language processing. In this work, I utilized Prolog along with Definite Clause Grammars (DCG) \cite{pereira1980definite} to solve one of the critical aspects of English grammar, active sentence and passive sentence. DCG proves the efficiency in handling the grammar of the sentence. Basically, a sentence is built out of noun phrase and verb phrase, so the structure of sentence, noun phrase, and verb phrase will be both covered in this work.

\begin{figure*}[ht]
\begin{subfigure}{.48\textwidth}
  \centering
  \includegraphics[width=0.9\linewidth]{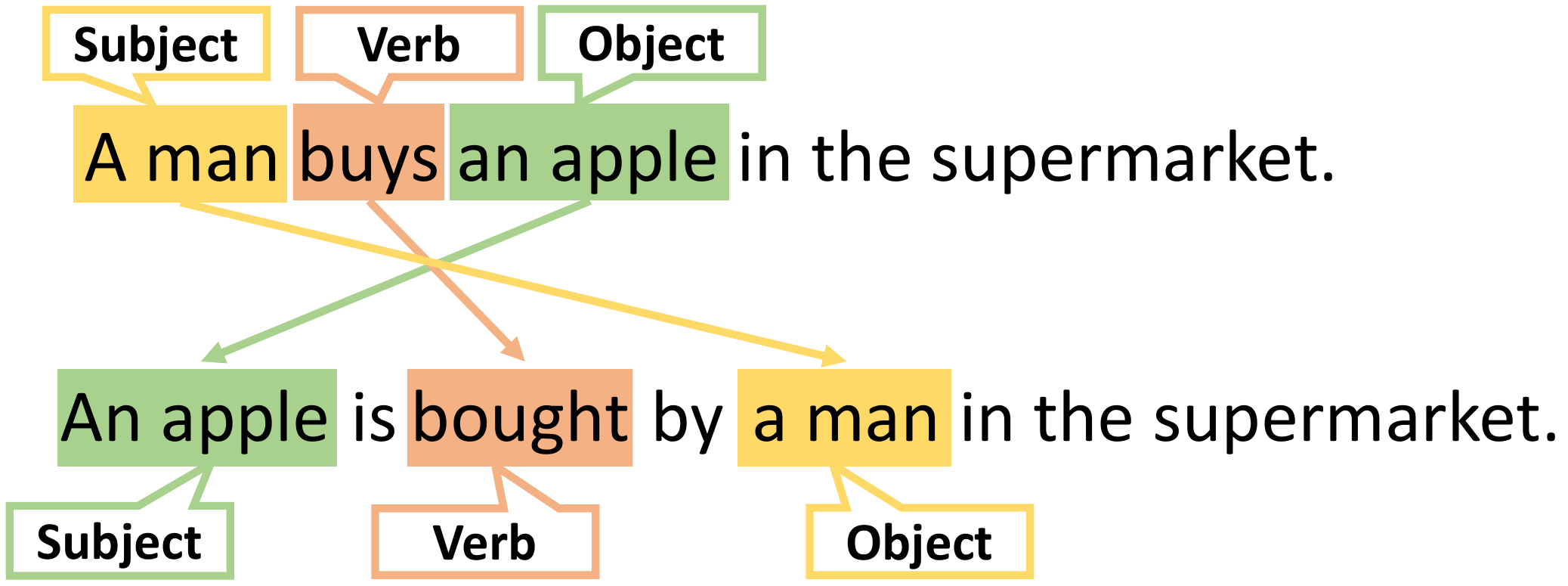}  
  \caption{The basic example}
  \label{fig:example1}
\end{subfigure}
\begin{subfigure}{.5\textwidth}
  \centering
  \includegraphics[width=0.88\linewidth]{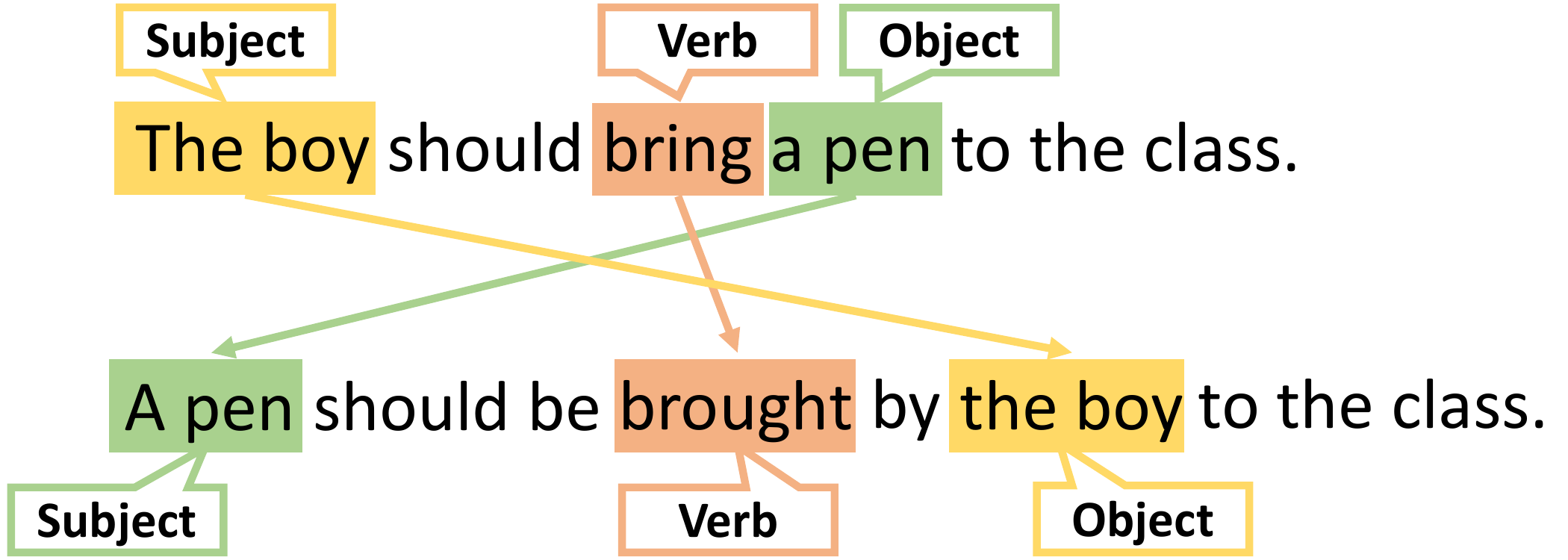}  
  \caption{The example of modal verb}
  \label{fig:example2}
\end{subfigure}
\caption{Basic rules for converting an active sentence to passive sentence}
\label{fig:example}
\end{figure*}

In terms of English grammar, we have lots of content to solve as shown in Figure \ref{fig:englishgrammar}. For example, there are 12 tenses in English such as the simple past tense, the simple present tense, the perfect present tense, etc. We also have more than three types of conditional clause, more than three types of comparative clause, and so on. This work covers the contents of active sentence and passive sentence. For instance, if an active sentence is \textit{``a man buys an apple in the supermarket"}, its corresponding passive sentence will be \textit{``an apple is bought by a man in the supermarket"}. The basic rules for rewriting an active sentence to
passive sentence are shown clearly in Figure \ref{fig:example}.

As shown in Figure \ref{fig:example}, basic rules are:

\begin{itemize}
    \item The object of the active sentence becomes the subject of the passive sentence;
    \item The subject of the active sentence becomes the object of the passive sentence;
    \item The finite form of the verb is changed to \textit{``to be + past participle"}.
\end{itemize}

As my best understanding so far, there are only a few works mentioning the problem of active sentence and passive sentence in terms of language processing and computational linguistics. The conversion between active sentence and passive sentence was early mentioned in \cite{stobo2004problem} by using a transformation rule to express the relationship between active and passive sentences. According to this rule, a parse tree is produced to represent the deep structure and determine whether the given sentence is active or passive. Similarly, \cite{nugues2006introduction} also used a tree-to-tree mapping to represent the active/passive transformation rule. However, these works just stopped in introducing how to transform an active sentence to passive sentence and did not solve many cases of them. Actually, there are many cases of active and passive sentences, leading to extra rules for converting between them. It is not easy to handle all these cases, and this is the main challenge of this work. My contributions are shown as follows:

\begin{enumerate}
    \item As far as I know, this may be the first work utilizing Prolog and DCG to solve a variety of cases of converting between active sentence and passive sentence such as 12 English tenses, modal verbs, negative form, etc.
    \item I proposed a \textbf{compact version} of the representation of the sentence as shown in Figure \ref{fig:compactrepresentation} and Figure \ref{fig:passiverepresentation}.
    \item In order to deal with 12 tenses in English, I proposed an \textbf{auxiliary-based solution} (is presented in Section \ref{auxiliarybasedsolution}) for dividing 12 tenses into 4 groups. This is a very nice solution that reduces the workload of defining DCG rules.
    \item I also proposed a \textbf{three-steps conversion} (is presented in Section \ref{threestepsconversion}) for doing the conversion between active sentence and passive sentence.
\end{enumerate}

\section{Analysis and Discussion}
\subsection{Cases to be solved}
The main challenge of this work is how much it can handle cases. There are a variety of cases in terms of active sentence and passive sentence. The cases that I solved in this work are shown as follows.

\begin{enumerate}
    \item The possibility of the conversion: the prerequisite to convert an active sentence to a passive sentence is that the active sentence must have the object. For instance:
    \begin{itemize}
        \item The sentence \textit{``the man buys an apple"} is converted to the passive form being \textit{``an apple is bought by the man"};
        \item However, the sentence \textit{``the man goes to school"} cannot be converted to the passive form because of the lack of object.
    \end{itemize}
    \item The tenses of the sentence: there are 12 tenses in English such as simple present tense, continuous past tense, perfect present tense, perfect continuous future tense, etc. With each tense, there is a \textbf{specific way} for converting between active sentence and passive sentence. For example (from active form to passive form):
    \begin{itemize}
        \item In the simple present tense: \textit{``the man buys an apple"} is converted to \textit{``an apple is bought by the man"};
        \item In the perfect present tense: \textit{``the man has bought an apple"} is converted to \textit{``an apple has been bought by the man"}.
    \end{itemize}
    This work handles all these 12 tenses.
    \item The form of past participle: commonly, a verb is converted to past participle form by adding \textit{``ed"} at the end (example: \textit{``add"} becomes \textit{``added"}, \textit{``look"} becomes \textit{``looked"}). However, there are some exceptions such as \textit{``buy"} becomes \textit{``bought"}, \textit{``see"} becomes \textit{``seen"}, etc.
    \item The case of negative sentence. For example, the negative form of \textit{``the man \textbf{buys} an apple"} is \textit{``the man \textbf{does not buy} an apple"}, and the corresponding passive sentence is \textit{``an apple \textbf{is not bought} by the man"}.
    \item The case of modal verb: modal verbs (also called modals, modal auxiliary verbs, modal auxiliaries) are special verbs which behave irregularly in English. They are different from normal verbs like \textit{``work"}, \textit{``play"}, \textit{``visit"}, etc. Modal verbs are always followed by an infinitive without \textit{``to"}. For example, the sentence \textit{``the boy \textbf{should bring} a pen to the class"} is converted to the passive form being \textit{``a pen \textbf{should be brought} by the boy to the class"} (Figure \ref{fig:example2}).
    \item Moreover, this work also handles the cases of singular/plural, subject pronoun/object pronoun, etc. For instance, the pronoun \textit{``he"} is used for the subject as \textit{``he"} but is used for the object as \textit{``him"}.
\end{enumerate}

\subsection{Representation and Inference}
The objective of this work is sentences: active sentence and passive sentence, so I need to determine the representation of both active sentence and passive sentence.

\begin{enumerate}
    \item An active sentence is built out of a noun phrase and a verb phrase. Therefore basically, the representation of an active sentence is \texttt{s(NP,VP)}. The noun phrase or verb phrase is built out of fundamental elements such as determiner, noun, adjective, verb, etc. Simply, the representation of fundamental elements are shown as follows:
    \begin{itemize}
        \item Determiner: \texttt{det(X)}. Example: \texttt{det(a)}, \texttt{det(an)}, \texttt{det(the)}, etc.
        \item Noun: \texttt{n(X)}. Example: \texttt{n(man)}, \texttt{n(woman)}, \texttt{n(apple)}, etc.
        \item Pronoun: \texttt{pro(X)}. Example: \texttt{pro(he)}, \texttt{pro(she)}, \texttt{pro(him)}, etc.
        \item Adjective: \texttt{adj(X)}. Example: \texttt{adj(small)}, \texttt{adj(big)}, \texttt{adj(beautiful)}, etc.
        \item Verb: \texttt{v(X)}. Example: \texttt{v(play)}, \texttt{v(like)}, \texttt{v(love)}, etc.
        \item Preposition: \texttt{pre(X)}. Example: \texttt{pre(on)}, \texttt{pre(in)}, \texttt{pre(by)}, etc.
        \item Auxiliary verb: \texttt{aux(X)}. Example: \texttt{aux(do)}, \texttt{aux(does)}, \texttt{aux(is)}, \texttt{aux(be)}, etc. Actually, there are three types of auxiliary verbs are used in this work. For example, the sentence \textit{``you \textbf{will have been} loving them"} (perfect continuous future tense) has three auxiliary verbs are \textit{``will"}, \textit{``have"}, \textit{``been"} which are determined by three predicates \texttt{aux/5}, \texttt{aux1/4}, \texttt{aux2/4} as shown in the source code (\textbf{convertible.pl}), respectively.
        \item Auxiliary verb for tense in the passive form: \texttt{auxTense(X)}. There are three groups of \texttt{auxTense}:
        \begin{itemize}
            \item Group 1: including only simple future tense: \texttt{auxTense(be)}. Example: \textit{``an apple will \textbf{be} bought buy the man"}.
            \item Group 2: consisting of continuous past tense, continuous present tense, continuous future tense, perfect continuous past tense, perfect continuous present tense, and perfect continuous future tense: \texttt{auxTense(being)}. Example: \textit{``an apple was \textbf{being} bought by a man"}, \textit{``an apple will be \textbf{being} bought by him"}.
            \item Group 3: including perfect past tense, perfect present tense, and perfect future tense: \texttt{auxTense(been)}. Example: \textit{``an apple has \textbf{been} bought by the man"}, \textit{``an apple will have \textbf{been} bought by the man"}.
        \end{itemize}
        \item Modal verb: \texttt{modal(X)}. Example: \texttt{modal(should)}, \texttt{modal(can)}, \texttt{modal(may)}, etc.
        \item Moreover, this work also uses \texttt{pol(not)} for the negative form and \texttt{agent(by)} for the passive form.
    \end{itemize}
    \item With a noun phrase, there are some ways to build the noun phrase such as:
    \begin{itemize}
        \item A noun phrase is built out of a determiner and a noun, so its representation is \texttt{np(DET,N)}. Example: noun phrase \textit{``the man"} has the representation is \texttt{np(det(the),n(man))}.
        \item A noun phrase is built out of pronoun such as \textit{``he"}, \textit{``she"}, \textit{``we"}, etc. In this case, the representation of the noun phrase is simply \texttt{np(PRO)}. For example: \texttt{np(pro(he))}.
        \item A noun phrase is built out of a determiner, adjectives, and a noun. In this case, the representation of the noun phrase is \texttt{np(DET,ADJ,N)}. For example, the noun phrase \textit{``a small beautiful girl"} has the representation is \texttt{np(det(a),adi([small, beautiful]), n(girl))}.
        \item A noun phrase is built out of a noun phrase and a prepositional phrase. The representation of the noun phrase in this case is \texttt{np(DET,N,PP)}, \texttt{np(PRO,PP)}, or \texttt{np(DET,ADJ,N,PP)}. For example, the noun phrase \textit{``a cat on the big table"} has the representation is
        
        \texttt{np(det(a),n(cat),pp(pre(on),det(the),adj([big]),n(table)))}.
    \end{itemize}
    \item With a verb phrase, there are two ways to build the verb phrase:
    \begin{itemize}
        \item A verb phrase is built out of a verb and a noun phrase. In this case, the presentation of the verb phrase is \texttt{vp(V,NP)}. For example, the verb phrase \textit{``love a beautiful woman"} has the representation is \texttt{vp(v(love), np(det(a), adj([beautiful]), n(woman)))}.
        \item A verb phrase is built out of only a verb, so its representation is simply \texttt{vp(V)}. Example: \texttt{vp(v(love))} or \texttt{vp(v(eat))}. In fact, as presented above, in order to be able to convert from an active sentence to a passive sentence, the active sentence has to have the object. Therefore, the case of verb phrase \texttt{vp(V)} will not be considered in this work.
    \end{itemize}
    \item After having the representation of noun phrase and verb phrase, the representation of the sentence could be obtained.
    \begin{itemize}
        \item Originally, the active sentence \textit{``he buys an apple"} has the representation is 
        
        \texttt{s(np(pro(he)),vp(v(buys),np(det(an),n(apple))))}. 
        
        However, as presented above, this work only considers the case of verb phrase \texttt{vp(V,NP)}, so I proposed a \textbf{compact version} of the representation of the sentence as shown in Figure \ref{fig:compactrepresentation}.
        
        \begin{figure}[h!]
        \centering
        \includegraphics[scale=0.6]{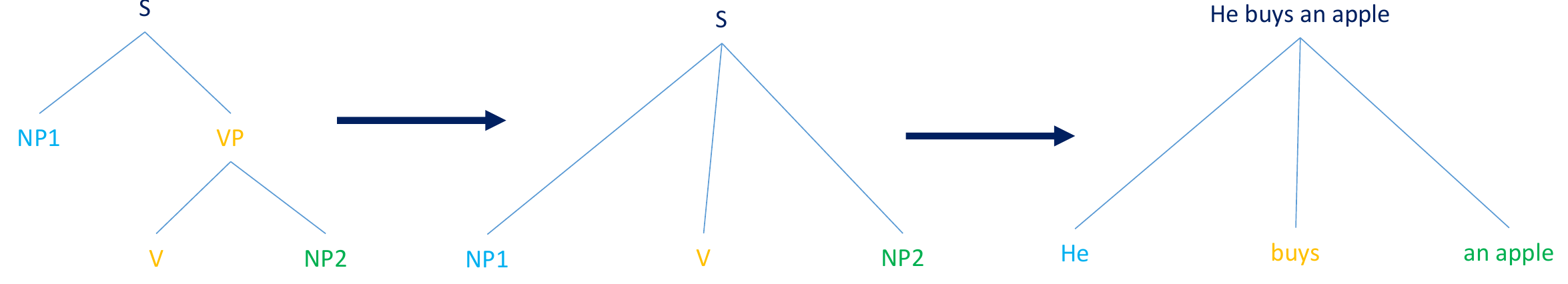}
        \caption{The compact version of the representation of the active sentence}
        \label{fig:compactrepresentation}
        \end{figure}
        
        Therefore, the active sentence \textit{``he buys an apple"} has the representation is 
        
        \texttt{s(np(pro(he)), v(buys), np(det(an), n(apple)))}.
        \item The passive sentence \textit{``an apple is bought by him"} has the representation is 
        
        \texttt{s(np(det(an), n(apple)), aux(is), v(bought), agent(by), np(pro(\newline him)))}.
    \end{itemize}
\end{enumerate}

\begin{figure}[h!]
\centering
\includegraphics[scale=0.6]{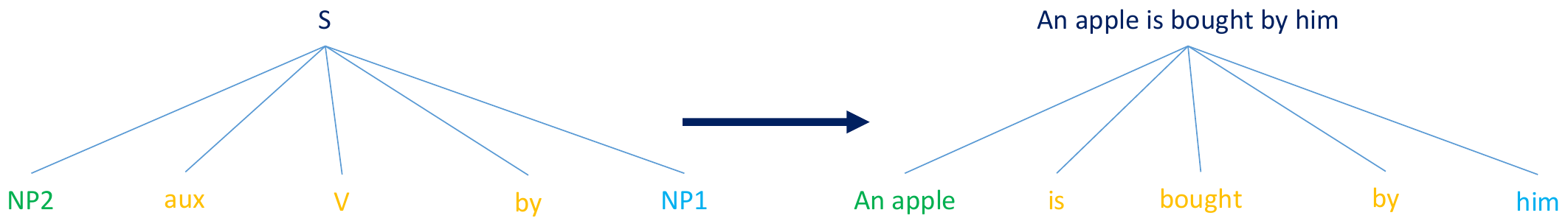}
\caption{The representation of the passive sentence}
\label{fig:passiverepresentation}
\end{figure}

As introduced in the DCG \cite{pereira1980definite}, the representation of the sentence is represented by \textbf{``parse tree"} as illustrated in Figure \ref{fig:compactrepresentation} (active sentence) and Figure \ref{fig:passiverepresentation} (passive sentence). Parse tree could be found with the help of advanced techniques like extra arguments and extra goals.

\textbf{``Inference"} is the conversion between a sentence and its representation, or even the conversion between an active sentence and a passive sentence:

\begin{itemize}
    \item Given a sentence, ``inference" is the process of getting the representation of that sentence;
    \item Given a representation of a sentence, ``inference" is the process of getting that sentence.
\end{itemize}

The final purpose of this work is that:

\begin{itemize}
    \item Given an active sentence, we will get the respective passive sentence; and vice versa,
    \item Given a passive sentence, we will get the respective active sentence.
\end{itemize}

\section{Design and Implementation}
\subsection{Scenario for user interaction}

\begin{figure}[h!]
\centering
\includegraphics[scale=0.6]{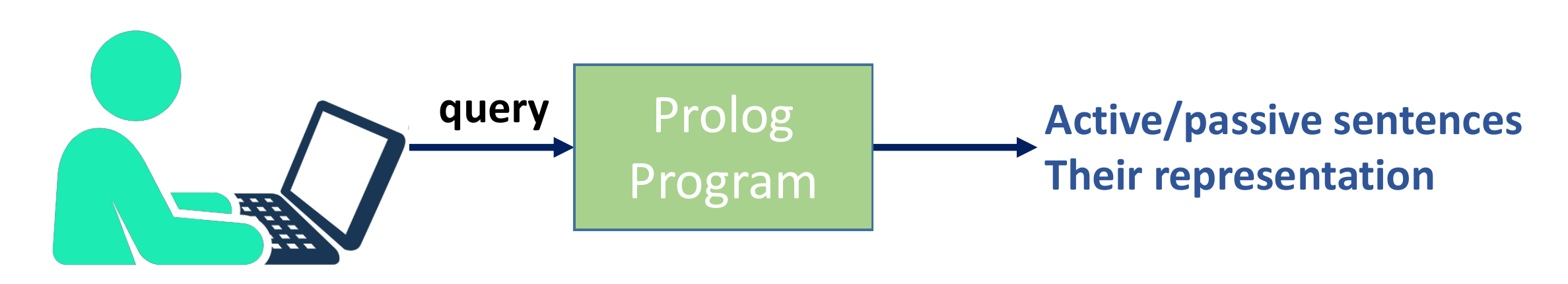}
\caption{The scenario for user interaction}
\label{fig:scenario}
\end{figure}

User interacts with the program by posing the query with the form (Figure \ref{fig:scenario}):

\begin{center}
\texttt{convert(ActiveS, ActiveRe, PassiveS, PassiveRe).}
\end{center}

Where:

\begin{itemize}
    \item \texttt{ActiveS}: the active sentence represented by a list where each element of the list corresponds to each word of the sentence. Example: \texttt{[he,buys,an,apple]}.
    \item \texttt{ActiveRe}: the representation of the active sentence \texttt{ActiveS}. 
    
    Example: \texttt{s(np(pro(he)),v(buys),np(det(an),n(apple)))}.
    
    \item \texttt{PassiveS}: the passive sentence represented by a list where each element of the list corresponds to each word of the sentence. Example: \texttt{[an,apple,is,bought,by,him]}.
    \item \texttt{PassiveRe}: the representation of the passive sentence \texttt{PassiveS}. Example:
    
    \texttt{s(np(det(an),n(apple)),aux(is),v(bought),agent(by),np(pro(him)))}.
    
\end{itemize}

\textbf{Input} will be either \texttt{ActiveS} or \texttt{PassiveS} for the case of converting from an active sentence to a passive sentence and the case of converting from a passive sentence to an active sentence, respectively.

There are several cases of \textbf{output}:

\begin{itemize}
    \item If the input is \texttt{ActiveS} and it is able to convert to the passive sentence, the outputs will be \texttt{ActiveRe}, \texttt{PassiveS}, and \texttt{PassiveRe}.
    \item If the input is \texttt{PassiveS} and it is able to convert to the active sentence, the outputs will be \texttt{ActiveS}, \texttt{ActiveRe}, and \texttt{PassiveRe}.
    \item If the input is either \texttt{ActiveS} or \texttt{PassiveS} but it is not able to convert to passive/active sentence, the output will be \texttt{‘false’}. There are some cases which cannot be converted:
    \begin{itemize}
        \item \texttt{ActiveS} is the active sentence but is typed as a passive sentence;
        \item \texttt{PassiveS} is the passive sentence but is typed as an active sentence;
        \item \texttt{ActiveS} is an active sentence having no object. Example: the sentence \textit{``he goes"} cannot be converted to the passive sentence.
    \end{itemize}
\end{itemize}

Especially, we can pose the query with no input, and the program will generate all possible cases of the active sentence and passive sentence. Some examples to make user interaction more clear will be presented in Section \ref{results}.

\subsection{Auxiliary-based solution to handle 12 English tenses}
\label{auxiliarybasedsolution}
\begin{figure}[h]
\centering
\includegraphics[scale=0.45]{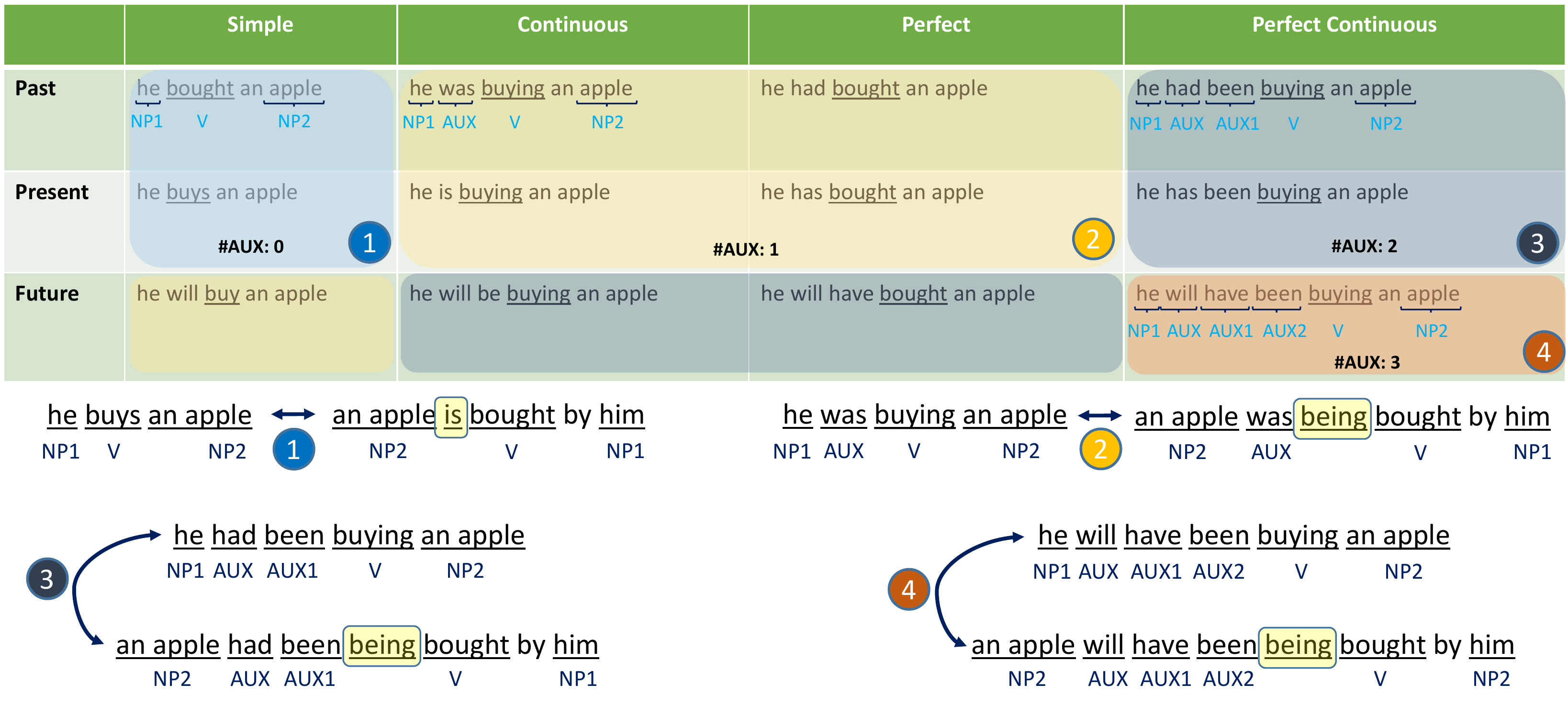}
\caption{Auxiliary-based solution: the division of 12 tenses into 4 groups}
\label{fig:tenses}
\end{figure}

There are 12 tenses in English. Each tense has a specific structure for the sentence. If each tense is handled individually, it will be quite long and be not an optimal solution. Therefore, as my best observation, I found a solution which divides 12 English tenses into 4 groups (same color means same group) based on the number of auxiliary verbs in the active sentence. This solution is summarized in Figure \ref{fig:tenses}, consisting of:

\begin{itemize}
    \item Group 1: the number of auxiliary verbs in the active sentence is equal to 0. This group consists of the simple past tense and the simple present tense;
    \item Group 2: the number of auxiliary verbs in the active sentence is equal to 1. We have 5 tenses in this group, those are the simple future tense, the continuous past tense, the continuous present tense, the perfect past tense, and the perfect present tense;
    \item Group 3: the number of auxiliary verbs in the active sentence is equal to 2. This group consists of the continuous future tense, the perfect future tense, the perfect continuous past tense, and the perfect continuous present tense;
    \item Group 4: the number of auxiliary verbs in the active sentence is equal to 3. This group has only one tense which is the perfect continuous future tense.
\end{itemize}

As we can easily see in Figure \ref{fig:tenses}, tenses in the \textbf{same group has the same structure of representation}. For example, DCG rules for active sentence and passive sentence of group 3 are implemented as follows.

\begin{lstlisting}
<@\textcolor{blue}{/*================== Group 3: For continuous\textunderscore future, perfect\textunderscore future, ==================*/}@>
<@\textcolor{blue}{/*=============== perfect\textunderscore continuous\textunderscore past, and perfect\textunderscore continuous\textunderscore present ===============*/}@>
<@\textcolor{blue}{/*============================ Positive - Active sentence =============================*/}@>
s(s(NP1,AUX,AUX1,V,NP2),Tense,Qs,Qo) -->
			np(NP1,subject,Qs),
			(
				{\+ NP1=np(pro(i))},
				aux(AUX,Tense,Qs)
			;
				{NP1=np(pro(i))},
				aux(AUX,Tense,special)
			),
			aux1(AUX1,Tense),
			v(V,Tense,Qs,group3),
			np(NP2,object,Qo).
			
<@\textcolor{blue}{/*============================ Positive - Passive sentence =============================*/}@>
s1(s(NP2,AUX,AUX1,AUX_TENSE,V,agent(by),NP1),Tense,Qs,Qo) -->
			np(NP2,subject,Qs),			
			(
				{\+ NP2=np(pro(i))},
				aux(AUX,Tense,Qs)
			;
				{NP2=np(pro(i))},
				aux(AUX,Tense,special)
			),
			aux1(AUX1,Tense),
			aux(AUX_TENSE,Tense),
			v(V,past_participle,Qs,past_participle),
			agent,
			np(NP1,object,Qo).		
\end{lstlisting}

\subsection{Three-steps conversion}
\label{threestepsconversion}
The three-steps conversion consists of three steps:

\begin{enumerate}
    \item From the input sentence fed as a list, the program first finds the representation of the sentence.
    \item From the representation of active or passive sentence, the program then finds the representation of passive or active sentence, respectively.
    \item From the representation achieved in the $2^{nd}$ step, the program returns the converted sentence as a list.
\end{enumerate}

The implementation of the three-steps conversion (written in \textbf{convert.pl}) is shown as follows.

\begin{lstlisting}
<@\textcolor{blue}{/*=============================== Three-steps conversion ===============================*/}@>
convert(ActiveS,ActiveRe,PassiveS,PassiveRe) :- 
	<@\textcolor{blue}{/*=== From Active sentence to Passive sentence ===*/}@>
	(\+ var(ActiveS);\+ var(ActiveRe)),
	s(ActiveRe,Tense,Qs,Qo,ActiveS,[]),		<@\textcolor{blue}{\% First step}@>
	write('Tense: '),write(Tense),write('\n'),
	convert(ActiveRe,Tense,Qs,Qo,PassiveRe),	<@\textcolor{blue}{\% Second step}@>	
	!,
	s1(PassiveRe,Tense,Qo,Qs,PassiveS,[])		<@\textcolor{blue}{\% Third step}@>	
;
	<@\textcolor{blue}{/*=== From Passive sentence to Active sentence ===*/}@>
	(\+ var(PassiveS);\+ var(PassiveRe)),
	s1(PassiveRe,Tense,Qo,Qs,PassiveS,[]),		<@\textcolor{blue}{\% First step}@>
	write('Tense: '),write(Tense),write('\n'),
	convert(ActiveRe,Tense,Qs,Qo,PassiveRe),	<@\textcolor{blue}{\% Second step}@>
	!,
	s(ActiveRe,Tense,Qs,Qo,ActiveS,[])		<@\textcolor{blue}{\% Third step}@>	
;
	<@\textcolor{blue}{/*=========== Generate all cases =================*/}@>
	var(ActiveS),var(ActiveRe),var(PassiveS),var(PassiveRe),
	s(ActiveRe,Tense,Qs,Qo,ActiveS,[]),		<@\textcolor{blue}{\% First step}@>
	write('Tense: '),write(Tense),write('\n'),
	convert(ActiveRe,Tense,Qs,Qo,PassiveRe),	<@\textcolor{blue}{\% Second step}@>	
	s1(PassiveRe,Tense,Qo,Qs,PassiveS,[]).		<@\textcolor{blue}{\% Third step}@>
\end{lstlisting}

The $1^{st}$ and $3^{rd}$ steps are done by using DCG rules (implemented in \textbf{convertible.pl}). The $2^{nd}$ step is easily done by the rule like:

\begin{lstlisting}
<@\textcolor{blue}{/*==================== Group 1: For simple\textunderscore present and simple\textunderscore past =====================*/}@>
<@\textcolor{blue}{/*================================== Positive sentence =================================*/}@>
convert(s(NP1,v(Y),NP2),Tense,Qs,Qo,s(NP22,aux(Aux),v(Y1),agent(by),NP11)) :-
	subAndObj(NP1,NP11),
	subAndObj(NP22,NP2),	
	(		
		\+ NP22=np(pro(i)),		
		lex(Aux,aux,Tense,Qo)	
	;			
		NP22=np(pro(i)),		
		lex(Aux,aux,Tense,special)
	),
	past_participle(Y,Y1,Qs,Tense).
\end{lstlisting}

As you can see above, the $2^{nd}$ step is easily done by doing the conversion between corresponding elements. More details for other groups are shown in \textbf{convert.pl}.

\subsection{Others}
All implementations above are for the positive form of the sentence. The negative form of the sentence can be easily done by inheriting the rules that are defined for the positive form. DCG rule for the negative form is implemented as follows.

\begin{lstlisting}
<@\textcolor{blue}{/*===================================== DCG rules ======================================*/}@>
<@\textcolor{blue}{/*========== Group 2: For simple\textunderscore future, continuous\textunderscore past, continuous\textunderscore present, ==========*/}@>
<@\textcolor{blue}{/*======================== perfect\textunderscore past, and perfect\textunderscore present ============================*/}@>
<@\textcolor{blue}{/*================================== Positive sentence =================================*/}@>
s(s(NP1,AUX,V,NP2),Tense,Qs,Qo) -->
			np(NP1,subject,Qs),
			(
				{\+ NP1=np(pro(i))},
				aux(AUX,Tense,Qs)
			;
				{NP1=np(pro(i))},
				aux(AUX,Tense,special)
			),
			v(V,Tense,Qs,group2),
			np(NP2,object,Qo).
			
<@\textcolor{blue}{/*================================== Negative sentence =================================*/}@>
s(s(NP1,AUX,<@\textbf{pol(not)}@>,V,NP2),Tense,Qs,Qo) -->
			np(NP1,subject,Qs),
			(
				{\+ NP1=np(pro(i))},
				aux(AUX,Tense,Qs)
			;
				{NP1=np(pro(i))},
				aux(AUX,Tense,special)
			),
			<@\textbf{pol}@>,
			v(V,Tense,Qs,group2),
			np(NP2,object,Qo).
\end{lstlisting}

DCG rules for the negative form is almost similar to those of the positive form, except from \texttt{pol/1} predicate. However, in the $2^{nd}$ step for the negative form, it completely utilizes the rule for the positive form as follows.

\begin{lstlisting}
<@\textcolor{blue}{/*===================================== $2^{nd}$ step =======================================*/}@>
<@\textcolor{blue}{/*========== Group 2: For simple\textunderscore future, continuous\textunderscore past, continuous\textunderscore present, ==========*/}@>
<@\textcolor{blue}{/*======================== perfect\textunderscore past, and perfect\textunderscore present ============================*/}@>
<@\textcolor{blue}{/*================================== Positive sentence =================================*/}@>
convert(s(NP1,aux(Aux1),v(Y),NP2),Tense,Qs,Qo,s(NP22,aux(Aux),auxTense(AuxTense),v(Y1),agent(by),NP11)) :-
	subAndObj(NP1,NP11),
	(		
		\+ NP1=np(pro(i)),		
		lex(Aux1,aux,Tense,Qs)	
	;			
		NP1=np(pro(i)),		
		lex(Aux1,aux,Tense,special)
	),	
	subAndObj(NP22,NP2),	
	(		
		\+ NP22=np(pro(i)),		
		lex(Aux,aux,Tense,Qo)	
	;			
		NP22=np(pro(i)),		
		lex(Aux,aux,Tense,special)
	),	
	lex(AuxTense,aux,Tense),	
	past_participle(Y,Y1,Qs,Tense).
	
<@\textcolor{blue}{/*================================== Negative sentence =================================*/}@>
convert(s(NP1,aux(Aux1),pol(not),v(Y),NP2),Tense,Qs,Qo,s(NP22,aux(Aux),pol(not),auxTense(AuxTense),v(Y1),agent(by),NP11)) :-
	convert(s(NP1,aux(Aux1),v(Y),NP2),Tense,Qs,Qo,s(NP22,aux(Aux),auxTense(AuxTense),v(Y1),agent(by),NP11)).
\end{lstlisting}

However, there is an exception of the $2^{nd}$ step for group 1, it needs an extra rule like:

\begin{lstlisting}
<@\textcolor{blue}{/*===================== Group 1: For simple\textunderscore present and simple\textunderscore past ====================*/}@>
<@\textcolor{blue}{/*================================== Positive sentence =================================*/}@>
convert(s(NP1,v(Y),NP2),Tense,Qs,Qo,s(NP22,aux(Aux),v(Y1),agent(by),NP11)) :-
	subAndObj(NP1,NP11),
	subAndObj(NP22,NP2),	
	(		
		\+ NP22=np(pro(i)),		
		lex(Aux,aux,Tense,Qo)	
	;			
		NP22=np(pro(i)),		
		lex(Aux,aux,Tense,special)
	),
	past_participle(Y,Y1,Qs,Tense).

<@\textcolor{blue}{/*================================== Negative sentence =================================*/}@>
convert(s(NP1,aux(AUX_POL),pol(not),v(Y),NP2),Tense,Qs,Qo,s(NP22,aux(Aux),pol(not),v(Y1),agent(by),NP11)) :-
	subAndObj(NP1,NP11),
	<@\textbf{lex(AUX\textunderscore POL,pol,Tense,Qs)}@>,
	subAndObj(NP22,NP2),	
	(		
		\+ NP22=np(pro(i)),		
		lex(Aux,aux,Tense,Qo)	
	;			
		NP22=np(pro(i)),		
		lex(Aux,aux,Tense,special)
	),
	past_participle(Y,Y1,plural,simple_present).
\end{lstlisting}

As we can see above, the negative form of group 1 needs the extra rule \texttt{lex(AUX\textunderscore POL,pol,Tense\newline ,Qs)} because, in this negative form, an extra auxiliary verb is needed. For example, the positive sentence is \textit{``he buys an apple"}, but the corresponding negative sentence is \textit{``he \textbf{does} not buy an apple"}. Other implementations such as lexicon, modal verbs, etc. are carefully written in the source code.

\section{Results}
\label{results}
This work has been already done with three files:

\begin{itemize}
    \item \textbf{convertible.pl}: implementing DCG rules for $1^{st}$ and $3^{rd}$ steps in the three-steps conversion, as well as other rules including lexicon.
    \item \textbf{convert.pl}: implementing the three-steps conversion and its $2^{nd}$ step.
    \item \textbf{testSuite.pl}: providing commands for user interaction. Users do not need to type the input sentence as a list (like \texttt{[the, man, buys, an, apple]}) but can type the sentence in the common way (directly type: \texttt{the man buys an apple}) by using two commands: \texttt{active} and \texttt{passive}. Moreover, users can easily check the correctness of the program by using two test suite commands: \texttt{activeTestSuite} and \texttt{passiveTestSuite}.
\end{itemize}

Some execution examples are shown as follows.

\begin{lstlisting}
?- active.
> a beautiful woman has bought a small apple on the big beautiful table.
ActiveS: [a,beautiful,woman,has,bought,a,small,apple,on,the,big,beautiful,table]
Tense: perfect_present
ActiveRe: s(np(det(a),adj([beautiful]),n(woman)),aux(has),v(bought),np(det(a),adj([small]),n(apple),pp(pre(on),np(det(the),adj([big,beautiful]),n(table)))))
PassiveS: [a,small,apple,on,the,big,beautiful,table,has,been,bought,by,a,beautiful,woman]
PassiveRe: s(np(det(a),adj([small]),n(apple),pp(pre(on),np(det(the),adj([big,beautiful]),n(table)))),aux(has),auxTense(been),v(bought),agent(by),np(det(a),adj([beautiful]),n(woman)))
true ;
false.

?- passive.
> a small apple should not be bought by him.
PassiveS: [a,small,apple,should,not,be,bought,by,him]
Tense: simple_present
ActiveS: [he,should,not,buy,a,small,apple]
ActiveRe: s(np(pro(he)),modal(should),pol(not),v(buy),np(det(a),adj([small]),n(apple)))
PassiveRe: s(np(det(a),adj([small]),n(apple)),modal(should),pol(not),aux(be),v(bought),agent(by),np(pro(him)))
true ;
false.
\end{lstlisting}

It should be noted that if users use \texttt{active} or \texttt{passive} commands, everything they type has to be defined in the lexicon or users have to define them in the lexicon (implemented in \textbf{convertible.pl}).

\section{Conclusion}
I introduced an effort to solve the problem of active and passive sentences using Prolog in terms of computation linguistics. By observing the possibility of converting an active sentence to passive sentence, I proposed a compact version of the representation of the sentence (Figure \ref{fig:compactrepresentation} and Figure \ref{fig:passiverepresentation}). I also introduced a solution called auxiliary-based solution (Section \ref{auxiliarybasedsolution}) to deal with 12 tenses in English. The auxiliary-based solution helps to reduce the workload of defining DCG rules. Finally, I proposed the three-steps conversion (Section \ref{threestepsconversion}) for converting between active sentence and passive sentence. In the future, this work should consider solving other cases of active and passive sentences as much as possible.

\bibliographystyle{plain}
\bibliography{references}

\end{document}